\documentclass[twoside]{article}
\usepackage{aistats2017}
  
\usepackage[utf8]{inputenc} 
\usepackage[T1]{fontenc}     
\usepackage{hyperref}           
\usepackage{url}                    
\usepackage{booktabs}         
\usepackage{amsfonts}         
\usepackage{amsmath}
\usepackage{amssymb}
\usepackage{tabularx}
\usepackage{nicefrac}       
\usepackage{microtype}      

\usepackage{makeidx} 
\usepackage{mathtools}
\usepackage{times}
\usepackage{natbib}
\usepackage{algorithm}
\usepackage{algorithmic}
\usepackage{graphicx} 
\usepackage{subfigure} 
\graphicspath{{figures/}} 
\usepackage{epstopdf}
\usepackage{sidecap}
\usepackage{comment}

\usepackage[usenames]{color}

\usepackage{enumitem}
\usepackage{bbold}
\usepackage{multirow}
\usepackage{pbox}
\usepackage{array}
\newcolumntype{L}[1]{>{\raggedright\let\newline\\\arraybackslash\hspace{0pt}}m{#1}}
\newcolumntype{C}[1]{>{\centering\let\newline\\\arraybackslash\hspace{0pt}}m{#1}}
\newcolumntype{R}[1]{>{\raggedleft\let\newline\\\arraybackslash\hspace{0pt}}m{#1}}

\newtheorem{thm}{Theorem}
\newtheorem{prop}{Proposition}

\newtheorem{defn}{Definition}
\newtheorem{lem}{Lemma}
\newtheorem{rem}{Remark}

\AtBeginDocument{\renewcommand{\bibname}{References}}

\begin{document}

%

%

\twocolumn[

\aistatstitle{Differentially Private Dropout}

\aistatsauthor{Beyza Ermi\c{s} \And Ali Taylan Cemgil }

\aistatsaddress{ Department of Computer Engineering \\ Bo\u{g}azi\c ci University, Istanbul, Turkey \And Department of Computer Engineering \\ Bo\u{g}azi\c ci University, Istanbul, Turkey} ]

\begin{abstract}
Large data collections required for the training of neural networks often contain sensitive information such as the medical histories of patients, and the privacy of the training data must be preserved. In this paper, we introduce a dropout technique that provides an elegant Bayesian interpretation to dropout, and show that the intrinsic noise added, with the primary goal of regularization, can be exploited to obtain a degree of differential privacy. The iterative nature of training neural networks presents a challenge for privacy-preserving estimation since multiple iterations increase the amount of noise added. We overcome this by using a relaxed notion of differential privacy, called concentrated differential privacy, which provides tighter estimates on the overall privacy loss. We demonstrate the accuracy of our privacy-preserving dropout algorithm on benchmark datasets.
\end{abstract}

\section{Introduction}
\label{sec:intro}
Deep neural networks (DNN) have recently generated significant interest, largely due to their successes in several important learning applications, including image classification, language modeling and many more (e.g.,~\citep{Lecun98,MaddisonHSS14,NIPS2015_5635}). 
The success of neural networks is directly related to the availability of large and representative datasets for training. However, these datasets are often collected from individuals, such as their tastes and behavior as well as medical health records, and present obvious privacy issues. Their usage requires methods that provide precise privacy guarantees while meeting the demands of the applications.

Overfitting is a key challenge in training deep neural networks, since DNNs can model highly complex prediction functions using a large number of parameters. 
It is often difficult to optimize these functions due to the potentially large number of local minimas in the space of parameters, and standard optimization techniques are prone to getting stuck in a local minimum which might be far from the global optimum. A popular regularization technique to avoid such local minima is dropout~\citep{Hinton2012,wang2013fast,Srivastava2014} which introduces noise into a model and optimizes loss function. 
Recently, it was shown that dropout can be treated as a Bayesian regularization method~\citep{kingma2015variational,Gal2015DropoutB} and it can be used to tune each weight's individual dropout rates. 
Besides its primary objective of regularization, dropout can be used to hide the details of the training data for achieving privacy. 
The key purpose of this paper is to analyze dropout in order to provide a theoretical guarantee for the privacy protection of the deep neural networks.

We develop a differentially private dropout technique by exploiting the inherent randomization of the dropout. Differential privacy (DP) is currently a widely accepted privacy definition~\citep{differential-privacy} to formalize the privacy protection of algorithms. The main principle of DP is to ensure that an adversary should not be able to reliably infer whether or not a particular individual is participating in a database, even with unlimited computational power and access to every entry except for that particular individual's data. 
This can be accomplished through adding noise into an algorithm at different stages such as adding noise to data itself or changing the objective function to be optimized. 
In order to design efficient differentially private algorithms, one needs to design a noise injection mechanism such that there is a good trade-off between privacy and utility. 
However, iterative algorithms such as stochastic gradient descent (SGD) accumulate the privacy loss at each iteration and a large number of iterations cause high cumulative privacy loss because of the potential of each access to leak more information. Therefore, we employ zCDP composition analysis~\citep{bun2016concentrated} that is inspired by concentrated differential privacy (CDP)~\citep{dwork2016concentrated} which is a recently proposed DP concept. CDP is well suited for iterative algorithms since it provides high probability bounds for cumulative privacy loss and requires adding much less noise for the same expected privacy guarantee compared to the DP.

In this paper, we study Gaussian dropout in the case we tune individual dropout rates for each weight of neural network to provide measurable privacy guarantee.  
Our main contributions can be summarized as follows:
\begin{itemize}
\item We first use the recently proposed connection between Gaussian dropout and Stochastic Gradient Langevin Dynamics (SGLD)~\citep{LiSCPGC16}, and then analyze that under what conditions dropout ensures to protect the privacy of the training data of DNNs.
\item In order to use the privacy budget more efficiently over many iterations, our approach uses the zCDP composition combined with the privacy amplification effect due to subsampling of data, which significantly decreases the amount of additive noise for the same expected privacy guarantee compared to the standard DP analysis. 
\item We empirically show that for general single hidden layer neural network models, dropout helps to regularize network and improves prediction accuracy while providing $(\epsilon,\delta)$-DP and zCDP. As our experiments illustrate, dropout with zCDP outperforms both the standard DP and the state-of-the-art algorithms, especially when the privacy budget is low. 
\end{itemize}

\section{Related Work}
\label{sec:related}
There are a number of works that address deep learning under differential privacy. Recently, \citep{ShokriS15} designed a system that enables multiple parties to train a neural network model without sharing their datasets. 
\citep{Phan0WD16} introduced a different approach towards differentially private deep learning that focuses on learning autoencoders by perturbing the objective functions of them. 
\citep{PapernotAEGT16} proposed a method where privacy-preserving models are learned locally from disjoint datasets and then combined in a privacy-preserving fashion. 
\citep{jain2015drop} used dropout to protect privacy in a completely different setting. They use binary dropout that cannot be converted to additive noise; so they add Laplace noise to the objective function to determine the dropping nodes from the network. Namely, they cannot use the intrinsic noise of the dropout. In addition, they use a different notion of DP, which is local-DP~\citep{duchi2013local}. 
Most recently, \citep{phan2017adaptive} developed a Laplace mechanism to preserve differential privacy in deep learning. This mechanism is independent of the number of training epochs in consuming privacy budget and it intentionally adds more (Laplace) noise into input features which are less relevant to the model output, and vice-versa.
 
Our work is most closely related to the study in \citep{Abadi2016}. In this work, they developed the moments accountant method, which is closely related to the notion of concentrated-DP, to accumulate the privacy cost that provides a tighter bound for privacy loss than previous composition methods. Then, by using the moments accountant they propose a differentially private SGD algorithm to train a neural network by perturbing the gradients of parameters in SGD. 
The gradients are perturbed by adding Gaussian distribution noise \emph{separately}. 
The addition of noise is a common technique for achieving privacy and also a common technique in deep learning~\citep{neelakantan2015adding,Abadi2016}. In our work, we show that there is a one to one relation between dropout rate and the Gaussian noise, so we say that dropout preserves ($\epsilon$, $\delta$)-DP in each DNN model for free when we calibrate the dropout rate carefully. Conversely, the amount of noise computed to provide ($\epsilon$, $\delta$)-DP also finds a proper dropout rate. 
We exploit the intrinsic randomized noise of the dropout by using two connections: i) Dropout and SGLD and ii) SGLD and DP. 
Besides, we apply the concentrated DP~\citep{dwork2016concentrated,bun2016concentrated}, which obtains equally tight bounds with the moments accountant but easier to convert into DP, to gradient perturbation mechanism. Then, we derive similar bounds to the moments accountant to strengthen the privacy guarantee in DNNs. Finally, we result in a similar algorithm with~\citep{Abadi2016}, but showing this similarity in the end is exactly the main goal of our paper. 

\section{Notations and Background}
\label{sec:backgr}
Throughout this paper, we assume that the data is $\mathcal{D} = \{d_i\}_{i=1}^N$ where $d_i$ = ($x_i$, $y_i$), with input object/feature $x_i \in \mathcal{R}^D$ and output label $y_i \in \mathcal{Y}$, with $\mathcal{Y}$ being the output discrete label space. A model characterizes the relationship from $x$ to $y$ with parameters (or weights) $\theta$.
Our goal is to tune the parameters $\theta$ of a model $p(y | x,\theta)$ that predicts $y$ given $x$ and $\theta$.
Bayesian inference in such a model consists of updating some initial belief over parameters $\theta$ in the form of a prior distribution $p(\theta)$, after observing data $\mathcal{D}$, into an updated belief over these parameters in the form of the posterior distribution $p(\theta | \mathcal{D})$.
The posterior distribution of a set of $N$ items is $p(\theta | \mathcal{D}) \propto p(\theta) \ p(\mathcal{D} | \theta)$ where the corresponding data likelihood is $p(\mathcal{D} | \theta) = \prod_{i=1}^N p(d_i | \theta)$.

Computing the posterior is often difficult in practice as it requires the computation of analytically intractable integrals, so we need to use approximation techniques. One of such techniques is Stochastic Gradient Langevin dynamics (SGLD)~\citep{WellingT11} that is used to scale up Bayesian learning by combining a popular class of methods called stochastic optimization~\citep{robbins1951} and Markov chain Monte Carlo~\citep{Robert2005} that generates a sequence of samples from a Markov chain.

\subsection{SGLD}
\label{sec:sgld}
Stochastic sampling methods such as SGLD~\citep{WellingT11} incorporate uncertainty into predictive estimates by running a perturbed version of the minibatch stochastic gradient descent on the negative log-posterior objective function: 
\small
\begin{align*}
\mathcal{L}(\theta) =  - \log p(\theta) - \frac{N}{S}\sum_{i \in \mathcal{S}_t} \log p(d_{t_i} | \theta)    
\end{align*}
\normalsize

At each iteration $t$, a subset of $S < N$ data items $\mathcal{S}_t = \{d_{t_1}, \cdots,  d_{t_S}\}$ is given and a Gaussian noise is injected during parameter updates so that they do not collapse to just the MAP solution. The parameters are updated as:
\begin{align*}
\theta_{t+1} \leftarrow \theta_t + \frac{\eta_t}{2} \nabla\theta_t + \zeta_t \ \text{,} \qquad \zeta_t \sim \mathcal{N}(0, \eta_t\mathbb{I}) 
\end{align*}
where $\{\eta_t\}$ is a sequence of step sizes and $\mathbb{I}$ is the identity matrix. 
To ensure convergence to a local maximum, a major requirement is for the step sizes to satisfy the properties: $\sum_{t=1}^\infty \eta_t = \infty$ and $\sum_{t=1}^\infty \eta_t^2 < \infty$. 
Given a set of samples from the update rule, posterior distributions can be approximated via Monte Carlo approximations as $p(y | x, \mathcal{D}) \approx \frac{1}{T} \sum_{t=1}^T p(y | x, \theta_t)$, where $T$ is the number of samples.

\subsection{Differential Privacy}
\label{sec:dp}
A natural notion of privacy protection prevents inference about specific records by requiring a randomized query response mechanism that yields similar distributions on responses of similar datasets. Formally, for any two possible input datasets $\mathcal{D}$ and $\mathcal{D}^\prime$ with the edit distance or Hamming distance $d(\mathcal{D}, \mathcal{D}^\prime)=1$, and any subset of possible responses $R$, a randomized algorithm $\mathcal{A}$ satisfies ($\epsilon$, $\delta$) differential privacy if 
\small
\begin{align}
P(\mathcal{A}(\mathcal{D}) \in R) \leq e^{\epsilon} P(\mathcal{A}(\mathcal{D}^\prime) \in R) + \delta 
\label{eqn:dp}
\end{align}
\normalsize
($\epsilon$, $\delta$)-differential privacy ensures that for all adjacent $\mathcal{D}$, $\mathcal{D}^\prime$, the absolute value of the privacy loss will be bounded by $\epsilon$ with probability at least $1-\delta$. 
Here, $\epsilon$ controls the maximum amount of information gain about an individual's data given the output of the algorithm. When the positive parameter $\epsilon$ is smaller, the mechanism provides stronger privacy guarantee~\citep{differential-privacy}.

\subsection{Concentrated Differential Privacy (CDP)}
\label{sec:cdp}
CDP is a recent variation of differential privacy which is proposed to make privacy-preserving iterative algorithms more practical than  DP while still providing strong privacy guarantees.
The CDP framework treats the privacy loss of an outcome,
$L^{(o)}_{(\mathcal{A}(\mathcal{D})\parallel \mathcal{A}(\mathcal{D}^\prime))} = \log\frac{P(\mathcal{A}(\mathcal{D})=o)}{P(\mathcal{A}(\mathcal{D}^\prime)=o)}$
as a random variable.
Two CDP methods are proposed in the literature. The first one is ($\mu$, $\tau$)-mCDP~\citep{dwork2016concentrated} where $\mu$ is the mean of this privacy loss. After subtracting $\mu$ from the resulting random variable $L^{(o)}$ is subgaussian with standard deviation $\tau$, i.e. $\forall\lambda \in \mathbb{R}$: $E\left[e^{\lambda(L^{(o)}-\mu)}\right]\leq e^{\lambda^2\tau^2/2}$. 
The second one is $\tau$-zCDP~\citep{bun2016concentrated} and we first define the R\'enyi divergence between two probability distribution in order to define it. 
\begin{defn} \textbf{(R\'enyi Divergence): } Let $P_1$ and $P_2$ be probability distributions. For $\alpha \in (1, \infty)$, the R\'enyi Divergence is defined of order $\alpha$ of $P_1$ from $P_2$ as:
\small
\begin{align*}
D_\alpha(P_1 \parallel P_2) = \frac{1}{\alpha -1} \log\left(\mathop{\mathbb{E}}_{x \sim P_1} \left[ \left(\frac{P_1(x)}{P_2(x)} \right)^{(\alpha-1)}\right]\right)
\end{align*}
\label{defn:RD}
\end{defn}
\normalsize
$\tau$-zCDP~\citep{bun2016concentrated} arises from a connection between the moment generating function of $L(o)$ and the R\'enyi divergence between the distributions of $\mathcal{A}_{(\mathcal{D})}$ and $\mathcal{A}_{(\mathcal{D}^\prime)}$. We require: $e^{(\alpha-1)D_\alpha} = E\left[e^{(\alpha-1)L^{(o)}}\right] \leq e^{(\alpha-1)\alpha\tau}$, $\forall\alpha \in (1,\infty)$. Observe that in this case $L^{(o)}$ is also subgaussian but zero-mean.

The zCDP definition is a relaxation of mCDP. In particular, a $(\mu,\tau)$-mCDP mechanism is also $(\mu-\tau^2/2, \tau^2/2)$-zCDP, but the opposite is not correct. Besides, zCDP can be thought of as providing guarantees of $(\epsilon, \delta)$-DP for all values of $\delta > 0$ (It will be shown in detail in Section~\ref{sec:methods}).
Accordingly, most of the DP mechanisms and applications can be characterized in terms of zCDP, but not in terms of mCDP, so we use zCDP as a tool for analyzing composition under the ($\epsilon$, $\delta$)-DP privacy definition, for a fair comparison between CDP and DP analyses.

\subsection{Dropout}
\label{sec:vd}
Dropout is one of the most popular regularization techniques for neural networks which injects multiplicative random noise to the input of each layer during the training procedure. For a fully connected neural network, the formalization of dropout is denoted as:
\begin{align}
h_2 = w((\xi \circ \theta) h_1)  \quad  \text{with} \quad  \xi_{i,j} \sim p(\xi_{i,j})
\end{align}
where $h_1$ and $h_2$ are the consecutive layers, $\theta$ is the weight matrix for the current layer and $w(\cdot)$ is the nonlinear function. The $\odot$ symbol denotes the elementwise (Hadamard) product of the input matrix with a matrix of independent noise variables $\xi$. The previous publications~\citep{Hinton2012,wan2013regularization,Srivastava2014} show that the weight parameters $\theta$ are less likely to overfit to the training data by adding noise to the input or weights during optimization.
At first, Hinton \emph{et al.}~\citep{Hinton2012} proposed the Binary Dropout where the elements of $\xi$ are drawn from a Bernoulli distribution with parameter $1-p$, hence each element of the input matrix is put to zero with probability $p$ that is also known as \emph{dropout rate}. Afterwards, the same authors proposed the  Gaussian Dropout using continuous noise $\xi_{i,j} \sim \mathcal{N}(1,\alpha=\frac{p}{1-p})$ with same relative mean and variance works as well or better~\citep{Srivastava2014}. 

\section{Methodology}
\label{sec:methods}
We describe our approach toward differentially private training of neural networks and introduce the proposed differentially private dropout algorithm. 
We first define the connection between dropout and SGLD by replacing the multiplicative noise term of dropout with an additive noise term.  
Then, we present our dropout algorithm and compute the per-iteration privacy budget for both advanced composition and zCDP composition.

\subsection{Connection between dropout and SGLD}
\label{sec:dropout_sgld}
Dropout has been proposed to improve model generalization in neural networks by adding noise to the local units or global weights during training. In~\citep{LiSCPGC16}, Molchanov \emph{et al} proved that the Stochastic Gradient MCMC model learning provides a Bayesian interpretation for dropout~\citep{wan2013regularization}. 
We have the update rule that shares the same form as SGLD by combining the SGD update with Gaussian dropout:
\begin{align}
\theta_{t+1} & = \big(\xi \circ \theta_t\big) + \frac{\eta_t}{2} \nabla_\theta g   \label{eq:SGLD2}   \\
                      & = \theta_t + \frac{\eta_t}{2} \nabla_\theta g + \xi^\prime  \label{eq:SGLD3}
\end{align}
where $\xi^\prime \sim \mathcal{N}(0, \eta_t V)$ and $V=\frac{\alpha}{\eta_t} diag(\theta_t^2)$. At each step, the gradient $\nabla_\theta g = \nabla \mathcal{L}(\theta)$ is computed for a random subset of examples as: 
\begin{align}
\nabla_\theta g = \nabla_\theta \log p(\theta_t) + \frac{N}{S} \sum_{i \in \mathcal{S}_t} \nabla_\theta \log p(d_{t_i} | \theta_t)  
\end{align}

In this way, we replace the multiplicative noise term $\xi$ in Eq.(\ref{eq:SGLD2}) with an additive noise term $\xi^\prime$ in Eq.(\ref{eq:SGLD3}) and it can be interpreted as injected noise from the Brownian motion of Langevin dynamics. The additive noise also helps us to propose a dropout algorithm that satisfies the differential privacy definition by adding independent Gaussian noise to the updates of each weight.

\subsection{Differentially Private Dropout (DPD)}
\label{sec:dp_dropout}
Eq.(\ref{eq:SGLD3}) is used to learn and regularize the model by adding random noise $\xi^\prime$. Our idea is to tune the noise term to provide privacy protection while keeping an acceptable dropout rate $\alpha$ to improve the model. To protect the privacy of training data, we need to perturb the gradients with a Gaussian noise in each iteration. 
We use the existing random noise $\xi^\prime \sim \mathcal{N}(0, \eta_t V)$ that is already being used for regularization where $V=\frac{\alpha}{\eta_t} diag(\theta_t^2)$ and eventually it equals to $\xi^\prime \sim \mathcal{N}(0, \alpha \ diag(\theta_t^2))$. 

In the original method~\citep{LiSCPGC16}, the authors considered the case when there is a single $\alpha$ for the model and the noise is controlled by it. 
In our case, at each iteration of the training scheme, \emph{DPD} takes a minibatch of data and computes the gradient, clips the $L_2$ norm of each gradient and adds noise to the gradient to protect privacy. The noisy stochastic gradient is then used to update model parameters $\theta$ via SGLD method by iteratively applying the update equation (\ref{eq:SGLD3}). 
The algorithm requires several parameters to determine the privacy budget such as sampling frequency $\nu=S/N$ for subsampling within the dataset, a total number of iterations $T$ and clipping threshold $C$. Clipping the gradients using the threshold $C$ will lead $L_2$ sensitivity of gradient sum to be $2C$. 
We have chosen to perturb parameter updates with zero mean multivariate normal noise with covariance matrix $(2 C)^2 \sigma^2  \mathbb{I}$. 
Parameter $\sigma$ in noise level determines our total $\epsilon$ and the amount of noise is chosen to be equal to the $\xi^\prime \sim \mathcal{N}(0, 4 C^2 \sigma^2 \mathbb{I})$. This amount is equal to the $\xi^\prime \sim \mathcal{N}(0, \alpha \ diag(\theta_t^2))$, hence $\alpha$ can be obtained by using this equality after determining the noise level that preserves $\epsilon$-DP. Here, $\alpha$ is not fixed for each weight and is controlled by the noise level that protects $\epsilon$-DP.
Algorithm~\ref{alg:PrivateDropout} outlines our method for training a model with parameters $\theta$. 
In the next subsection, we describe in detail how privacy design parameters are chosen and privacy budget is calculated.
\small
\begin{algorithm}[h!]
\caption{Differentially Private Dropout (DPD)}
\begin{algorithmic}[1]
\STATE \textbf{Inputs}: Input data $\mathcal{D} = \{x_i, y_i\}_{i=1}^N$, number of data passes $T$, minibatch size $S$, learning rate $\eta_t$, noise scale $\sigma$, gradient norm bound $C$.
\STATE Initialize $\theta_0$ randomly
\FOR{ $t \leftarrow 1$ to $T$ }
\STATE Take a random sample $\mathcal{S}_t$ of size $S$ with sampling probability $\nu=S/N$ 
\STATE Compute gradient $\nabla_\theta g$ from $(x_i,y_i) \in \mathcal{S}_t$
\STATE Clip gradient: $\nabla_\theta \bar{g} = \nabla_\theta g/\max\big(1, {\parallel\nabla_\theta g\parallel_2}/{C} \big)$ 
\STATE Compute noise: $\xi^\prime \sim \mathcal{N}(0, 4 C^2 \sigma^2 \mathbb{I})$ 
\STATE Update parameter: $\theta_{t+1} = \theta_t + \frac{\eta_t}{2} \left(\nabla_\theta \bar{g} + \xi^\prime\right)$
\ENDFOR
\STATE \textbf{Output}: $\theta_T$.
\end{algorithmic}
\label{alg:PrivateDropout}
\end{algorithm}
\normalsize

\subsection{Per-iteration privacy budget}
\label{sec:dp_budget}
When developing a differentially private algorithm, we encounter a challenge that the number of iterations accumulates the privacy loss at each access to the training data, and the number of iterations required to guarantee accurate posterior estimates causes high cumulative privacy loss. 
In order to use the privacy budget more effectively across many iterations, one needs to calculate the privacy cost by using an enhanced composition analysis. 

In this work, we first calculate the per-iteration privacy budget using the key properties of advanced composition theorem (Theorem 3.20 of \citep{Dwork2014}) and this method is called \emph{DPD-AC} in the experiments. Then, we use a relaxed notion of differential privacy, called zCDP~\citep{bun2016concentrated} that bounds the moments of the privacy loss random variable and call this method \emph{DPD-zCDP}. The moments bound yields a tighter tail bound, and consequently, it allows for a higher per-iteration budget than standard DP-methods for a given total privacy budget.

\begin{thm} \textbf{(Advanced composition): } For all $\epsilon$, $\delta$, $\delta^\prime \geq$ 0, the class of ($\epsilon$, $\delta$)-DP mechanisms satisfy ($\epsilon_{tot}$, $\delta_{tot}$)-DP under k-fold adaptive composition for 
\begin{align*}
\epsilon_{tot}=\sqrt{2k\log(1/\delta^\prime)}\epsilon + k\epsilon(e^\epsilon-1)  \ ,  \quad  \delta_{tot} = k\delta + \delta^\prime
\end{align*}
\label{thm:AC}
\end{thm}

\begin{rem} \textbf{(Remark 1 in~\citep{icml2015_wangg15}): } When $\epsilon=\frac{c}{\sqrt{2k\log(1/\delta^\prime)}}<1$ for some constant $c<\sqrt{\log(1/\delta^\prime)}$, the equation of $\epsilon^\prime$ can be simplified into $\epsilon^\prime \leq 2c$ by applying the inequality $e^\epsilon -1\leq 2\epsilon$. 
\label{lem:rem1}
\end{rem}
The theorem states that with small $\epsilon$ and small loss in $\delta_{tot}$, more strict $\epsilon_{tot}$ is obtained than just summing the $\epsilon$. This is clear by looking at the first order expansion for small $\epsilon$ (Taylor Theorem is used with assumption that $\epsilon \leq 1$) of $\epsilon_{tot} = \sqrt{2 k \log(1/\delta^\prime)}\epsilon + k \epsilon^2$.

There are many ways to make an algorithm differentially private. In this paper, we add differential privacy into Gaussian dropout by clipping and perturbing the gradients. As our method for perturbation, we use a specific form of the global sensitivity method, called the \emph{Gaussian Mechanism}~\citep{differential-privacy}, where Gaussian noise calibrated to the global sensitivity is added. 

\begin{thm} \textbf{(Gaussian Mechanism (GM), in~\citep{Dwork2014}): } Let $\epsilon \in (0,1)$ be arbitrary. \emph{Gaussian Mechanism} states that given function $f$ with $L_2$ sensitivity of $\bigtriangleup_2 f$, releasing $f(X) + Z$ where $Z \sim \mathcal{N}(0,\sigma^2)$ is ($\epsilon, \delta$)-DP when $\sigma \geq \bigtriangleup_2 f \sqrt{2\log(1.25/\delta)}/\epsilon$.
\label{thm:GM}
\end{thm}
Given two adjacent datasets $\mathcal{D}$ and $\mathcal{D}^\prime$, the important $L_2$-sensitivity of a function $f$ is defined as:
\small
\begin{align*}
\bigtriangleup_2 f = \sup_{\mathcal{D},\mathcal{D}^\prime,\parallel \mathcal{D} - \mathcal{D}^\prime\parallel=1} \parallel f(\mathcal{D}) - f(\mathcal{D}^\prime) \parallel_2 
\end{align*}
\normalsize

We use a stochastic gradient algorithm that uses minibatches of data while learning, so we can make use of the \emph{amplifying effect of the subsampling} on privacy. 
The version of the privacy amplification theorem we use is as follows:
\begin{thm} \textbf{(Theorem 1 in~\citep{Li2012}): } Any ($\epsilon_{iter}$, $\delta_{iter}$)-DP mechanism running on a sampled subset of the data, where each data point is included independently with probability $\nu$, and where $\nu>\delta_{iter}$, guarantees ($\log(1+\nu(\exp(\epsilon_{iter})-1))$, $\nu \delta_{iter}$)-DP.
\label{thm:amplification}
\end{thm}

\noindent We will assume that the instances are included independently with probability $\nu= S/N$; and for ease of implementation, we will use minibatches with fixed size $S$ in our experiments. 
As we mentioned before, parameter $\sigma$ in noise level determines our total $\epsilon$ and depends on the total $\delta$ in privacy budget. We can calculate by the total privacy budget $\epsilon_{tot}$ by setting $\sigma = \sqrt{2\log(1.25/\delta_{iter})}/\epsilon_{iter}$.
Clipping will lead $L_2$ sensitivity of gradient sum to be $2 C$, so perturbing each sum with aforementioned noise will lead each iteration to be ($\epsilon_{iter}$,$\delta_{iter}$)-DP w.r.t the subset. Now if we set $\delta_{iter} = (\delta_{tot} - \delta^\prime)/T \nu$, where $\delta^\prime$ comes from advanced composition, we can provide $\delta_{tot}$ as $\delta$ parameter in total privacy cost. Using Theorem~\ref{thm:AC} and Theorem~\ref{thm:GM}, the $\epsilon$ parameter in our total privacy cost for \emph{DPD-AC} will be:
\begin{align*}
\epsilon_{tot} = \sqrt{2 T \log(1/\delta^\prime)}/\sigma^\prime + T (\sigma^\prime)^2  \quad  \text{where}  \\
\sigma^\prime = \log\big(1 + \nu \big(\exp (\sqrt{2 \log(1.25/\delta_{iter})}/\sigma) -1 \big) \big)
\end{align*} 

\paragraph{CDP:} For \emph{DPD-zCDP}, we calculate the per-iteration budget using \emph{the zCDP composition} that is used for tracking privacy loss of the composite mechanisms. It permits a sharper analysis of the per-iteration privacy budget. We first convert DP to zCDP, then use the zCDP composition. For comparison purposes, we convert zCDP back to DP at the end. We use the following lemmas and propositions for this process:

\begin{prop} \textbf{(Proposition 1.6 in~\citep{bun2016concentrated}): } The Gaussian mechanism with some noise variance $\tau$ and a sensitivity $\bigtriangleup$ satisfies $\bigtriangleup^2/(2\tau)$-zCDP.
\label{prop:Prop1.6}
\end{prop}

\begin{lem} \textbf{(Lemma 1.7 in~\citep{bun2016concentrated}): } If two mechanisms satisfy $\rho_1$-zCDP and $\rho_2$-zCDP, respectively, then their composition satisfies ($\rho_1+\rho_2$)-zCDP.
\label{lem:Lemma1.7}
\end{lem}

\begin{prop} \textbf{(Proposition 1.3 in~\citep{bun2016concentrated}): } If $\mathcal{A}$ provides $\rho$-zCDP, then $\mathcal{A}$ is ($\rho + 2 \sqrt{\rho \log(1/\delta)}, \delta$)-DP
\label{prop:Prop1.3}
\end{prop}
\noindent Using Proposition~\ref{prop:Prop1.6} and Lemma~\ref{lem:Lemma1.7}, we obtain $T\bigtriangleup^2/(2\tau)$-zCDP after $T$-composition of the Gaussian mechanism. Using Proposition~\ref{prop:Prop1.3}, we convert $T\bigtriangleup^2/(2\tau)$-zCDP to $(\rho+2\sqrt{\rho\log(1/\delta)},\delta)$-DP where $\rho=T\bigtriangleup^2/(2\tau)$.
We will summarize these seemingly complicated process under two straightforward steps: i) zCDP composition and ii) privacy amplification.

\noindent \textbf{zCDP composition:} Given a total privacy budget $\epsilon_{tot}$ and total tolerance level $\delta_{tot}$, our algorithm computes a privacy budget using the zCDP composition, which maps ($\epsilon_{tot}$, $\delta_{tot}$) to ($\epsilon^\prime$, $\delta^\prime$) as:
\small
\begin{align}
\epsilon_{tot} = J\bigtriangleup^2/(2\tau) & + 2\sqrt{J\bigtriangleup^2/(2\tau)\log(1/\delta_{tot})} , \label{eq:epsTot} \\ 
\text{where} \quad \tau & \geq 2\log(1.25/\delta^\prime)\bigtriangleup^2/{\epsilon^\prime}^2  \notag
\end{align}
\normalsize 

\noindent \textbf{Privacy amplification:} Our algorithm computes the per-iteration privacy budget ($\epsilon_{iter}$, $\delta_{iter}$) using the privacy amplification theorem, which maps ($\epsilon^\prime$,$\delta^\prime$) to ($\epsilon_{iter}$, $\delta_{iter}$):
\small
\begin{align}
\epsilon^\prime = \log(1+\nu(\exp(\epsilon_{iter})-1)) , \qquad  \delta^\prime = \nu \delta_{iter}
\label{eq:epsIter}
\end{align}
\normalsize

\section{Experiments and Results}
\label{sec:exp_res}
We evaluate our approach on two standard benchmark datasets: MNIST~\citep{Lecun98} that contains 70K $28 \times 28$ handwritten digits (60K for training and 10K for testing) and DIGITS~\citep{bache2013uci} that consists of 1797 $8 \times 8$ grayscale images (1439 for training and 360 for testing) of handwritten digits. 
We use a single layer feed-forward neural network with ReLU units and softmax of 10 classes for both datasets. We choose the non-private (NP) model as our baseline. 
For MNIST, we use the minibatch of size $S=600$ with 1000 hidden units and reach an accuracy of $98.20\%$ in about 200 epochs. For DIGITS, we use the minibatch of size $S=100$ with 500 hidden units and reach an accuracy of $95.50\%$ in about 100 epochs. 
All of the results are implemented in Theano~\citep{theano2016} and are the average of 10 runs.

\paragraph{Differentially private models:} We experiment with the same architecture for the differentially private version. To limit sensitivity, we clip the gradient norm of each layer at $C=3$ for MNIST and $C=2$ for DIGITS. We report results for three different noise levels where $\epsilon = \lbrace 10, 1, 0.5\rbrace$. 
For any fixed $\epsilon$, $\delta$ is varied between $10^{-2}$ and $10^{-5}$. There is a slight difference with different $\delta$ values (less than $10^{-3}$), but still we choose the best performing $\delta=10^{-4}$ for both datasets. We set the initial learning rate $\eta_0=0.1$ for MNIST and $\eta_0=0.05$ for DIGITS and update it per round as $\eta_t = \eta_0/t^\gamma$. We fixed the decay rate $\gamma$ to 1 for both of the datasets. 

\begin{figure*}[t]
\hspace*{-5mm}
\begin{minipage}[b]{0.33\textwidth}
\centering
\subfigure[DPD-AC (MNIST)]{\includegraphics[scale=0.25]{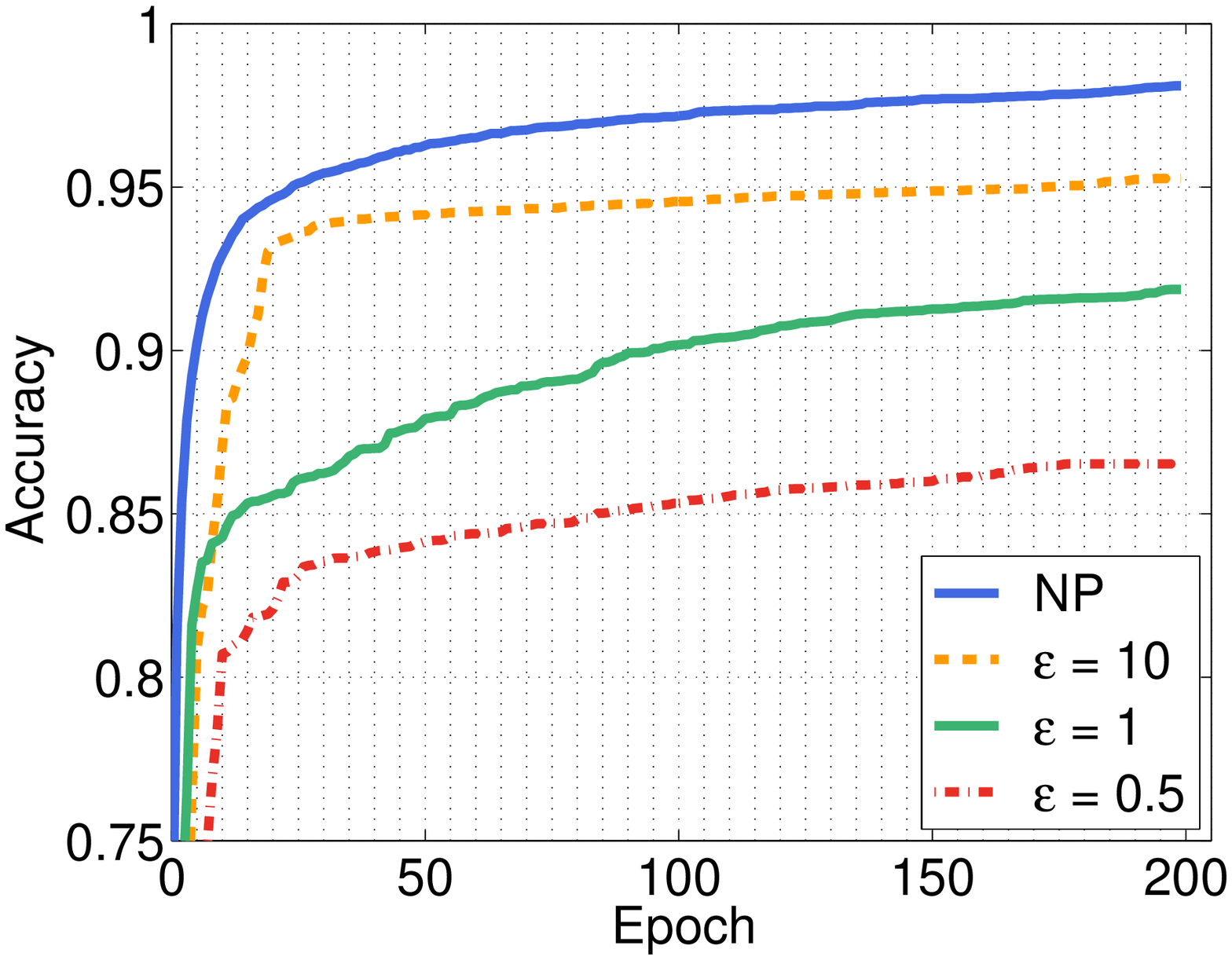} \label{subfig:DPmnist}}
\end{minipage}   
\hspace*{1mm}
\begin{minipage}[b]{0.33\textwidth} 
\centering
\subfigure[DPD-zCDP (MNIST)]{\includegraphics[scale=0.25]{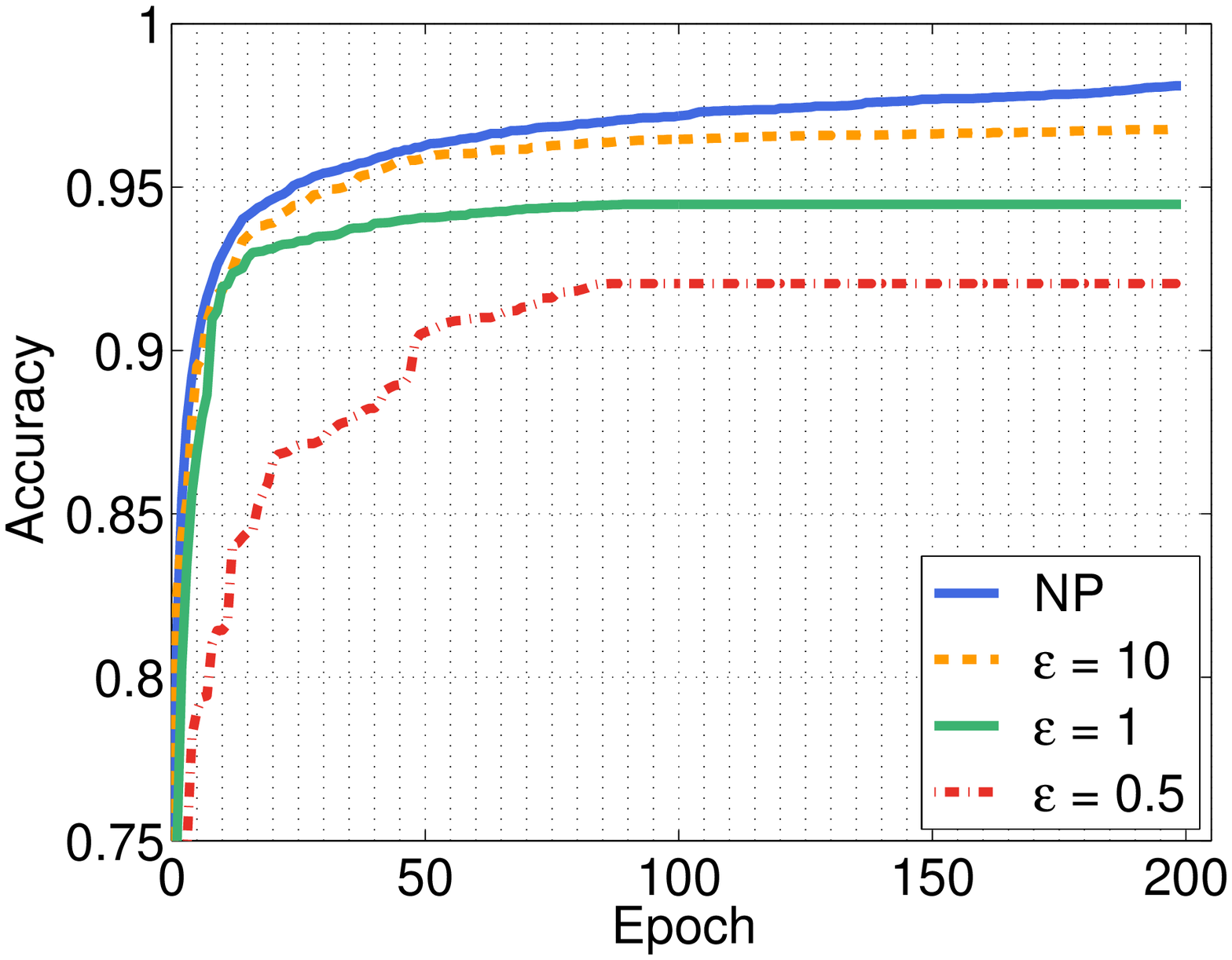} \label{subfig:zcDPmnist}}
\end{minipage}
\hspace*{1mm}
\begin{minipage}[b]{0.33\textwidth} 
\centering
\subfigure[$\sigma$ vs $\epsilon$ (MNIST)]{\includegraphics[scale=0.25]{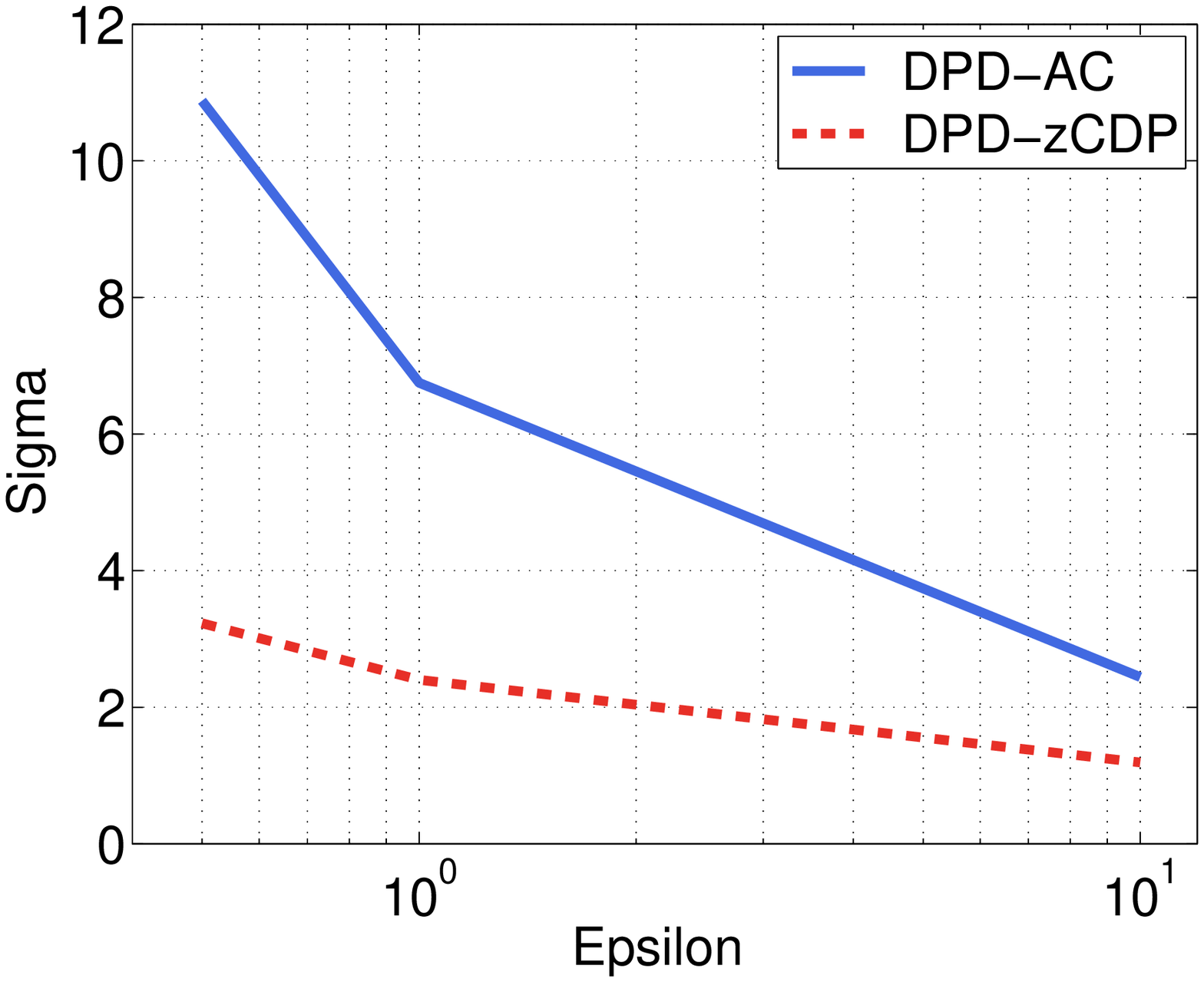} \label{subfig:compSigmaM}}
\end{minipage}
\\
\hspace*{-5mm}
\begin{minipage}[b]{0.33\textwidth}
\centering
\subfigure[DPD-AC (DIGITS)]{\includegraphics[scale=0.25]{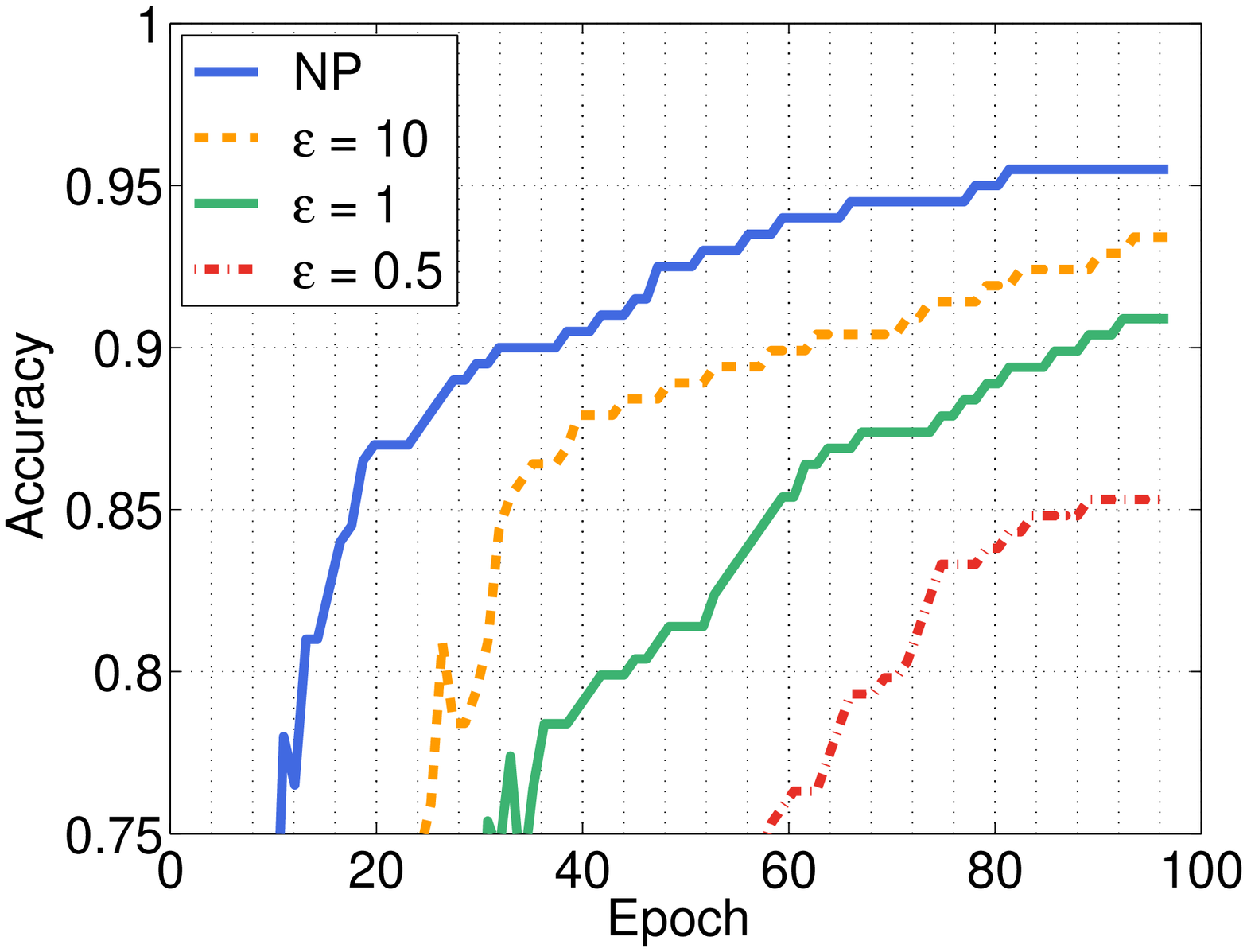} \label{subfig:DPdigits}}
\end{minipage}   
\hspace*{1mm}
\begin{minipage}[b]{0.33\textwidth} 
\centering
\subfigure[DPD-zCDP (DIGITS)]{\includegraphics[scale=0.25]{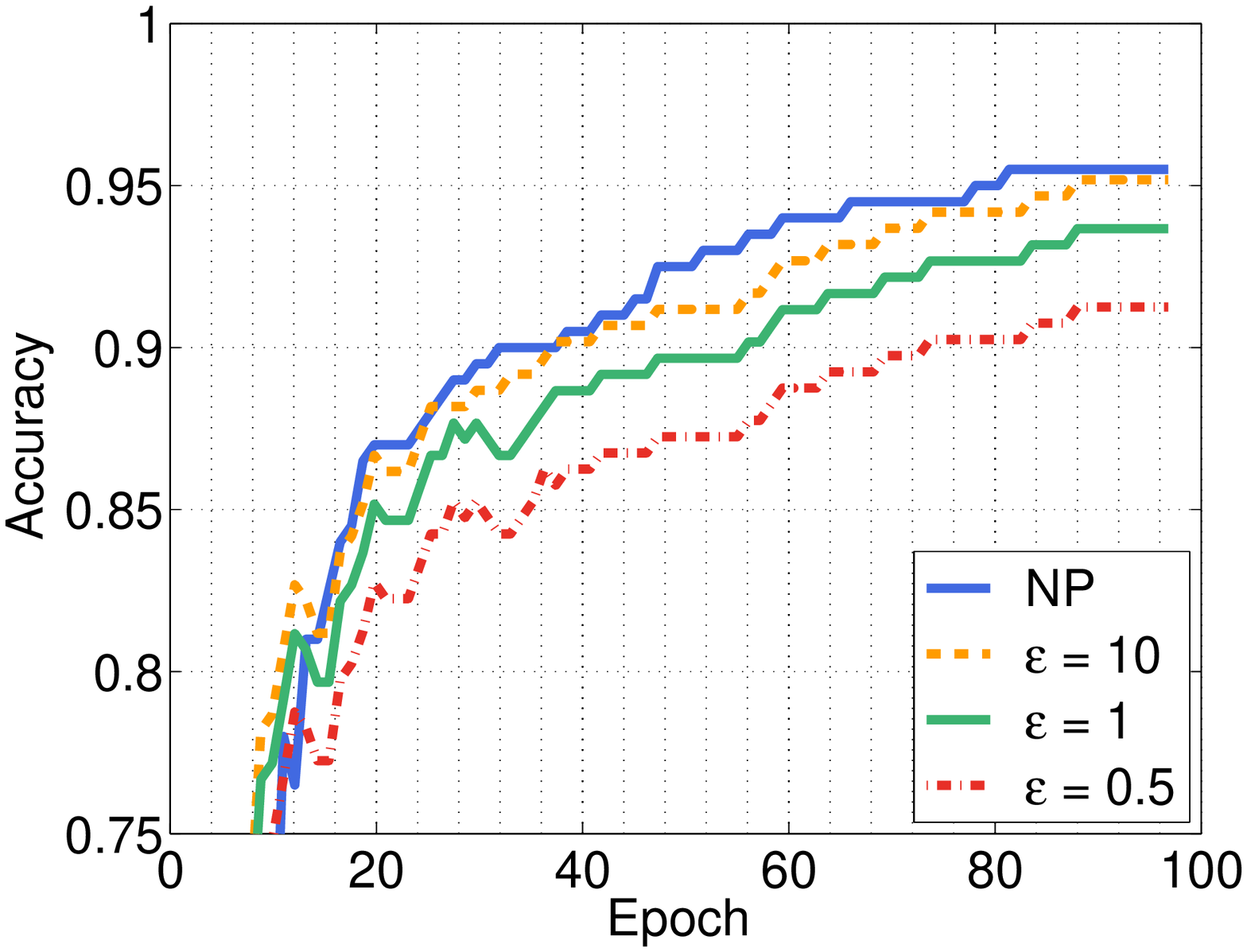} \label{subfig:zcDPdigits}}
\end{minipage}
\hspace*{1mm}
\begin{minipage}[b]{0.33\textwidth} 
\centering
\subfigure[$\sigma$ vs $\epsilon$ (DIGITS)]{\includegraphics[scale=0.25]{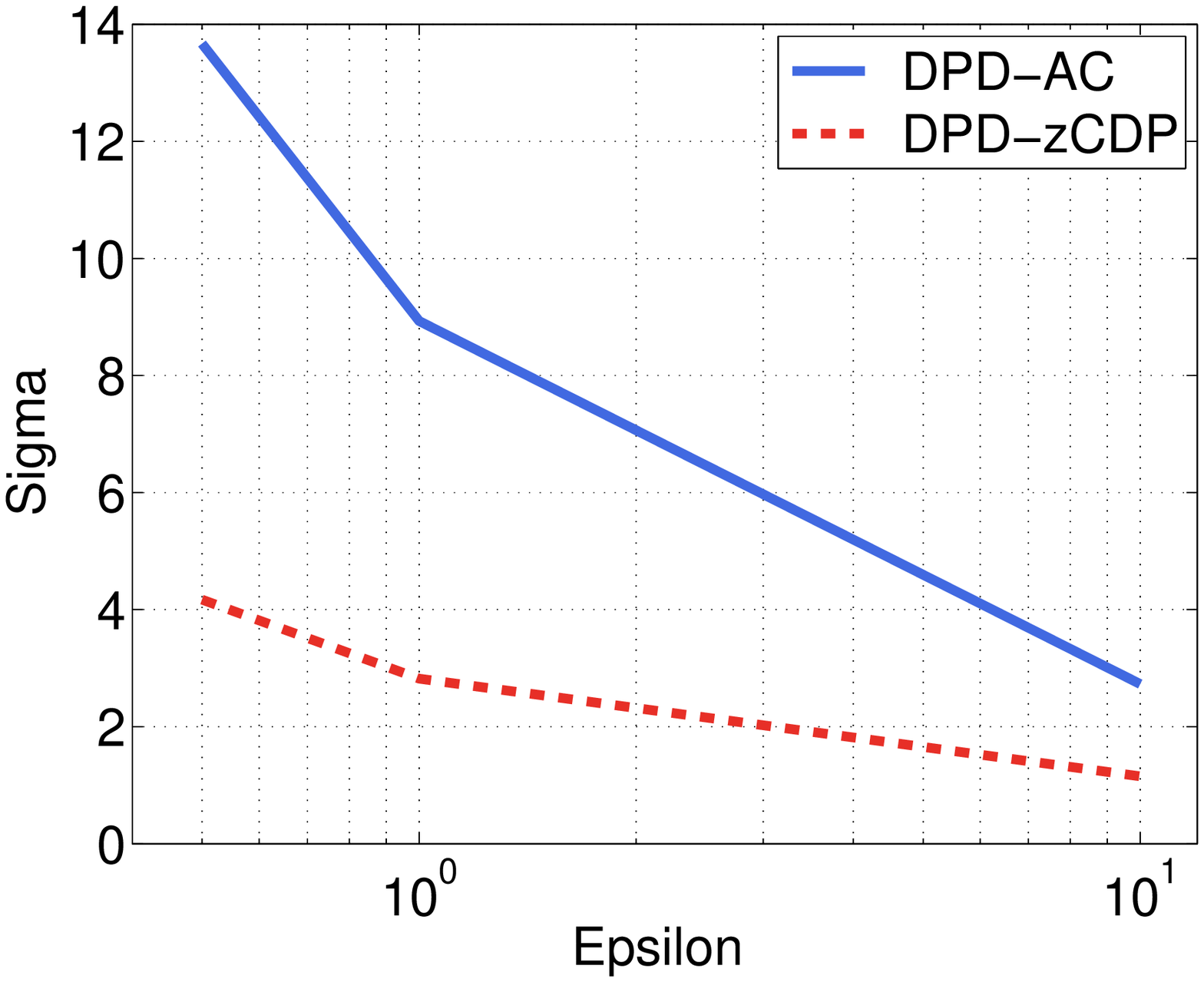} \label{subfig:compSigmaD}}
\end{minipage}
\caption{Comparison of the test accuracies for $\epsilon=\lbrace 10, 1, 0.5\rbrace$ and the NP case for (a) AC, b) zCDP compositions on MNIST and (d) AC, e) zCDP compositions on DIGITS. (c) and (f) shows the $\sigma$ value as a function of $\epsilon$ on MNIST and DIGITS respectively.}
\label{fig:Mnist}
\end{figure*}

In the first set of experiments, we investigate \emph{the influence of privacy loss on accuracy}. The mini-batch sizes were set to $S=600$ and $S=100$ for MNIST and DIGITS respectively. We ran the algorithm for 200 passes for MNIST and 100 passes for DIGITS. Figure~\ref{subfig:DPmnist},~\ref{subfig:DPdigits} report the performance of the DPD-AC method and Figure~\ref{subfig:zcDPmnist},~\ref{subfig:zcDPdigits} report the performance of the DPD-zCDP method for different noise levels. 
These results justify the theoretical claims that lower prediction accuracy is obtained when the privacy protection is increased by decreasing $\epsilon$. 
One more deduction from these results is that dropout with the zCDP composition (\emph{DPD-zCDP} method) can reach an accuracy very close to the non-private level especially under reasonably strong privacy guarantees (when $\epsilon > 0.5$).

\begin{figure*}[t]
\hspace*{-5mm}
\begin{minipage}[b]{0.33\textwidth} 
\centering
\subfigure[MNIST]{\includegraphics[scale=0.25]{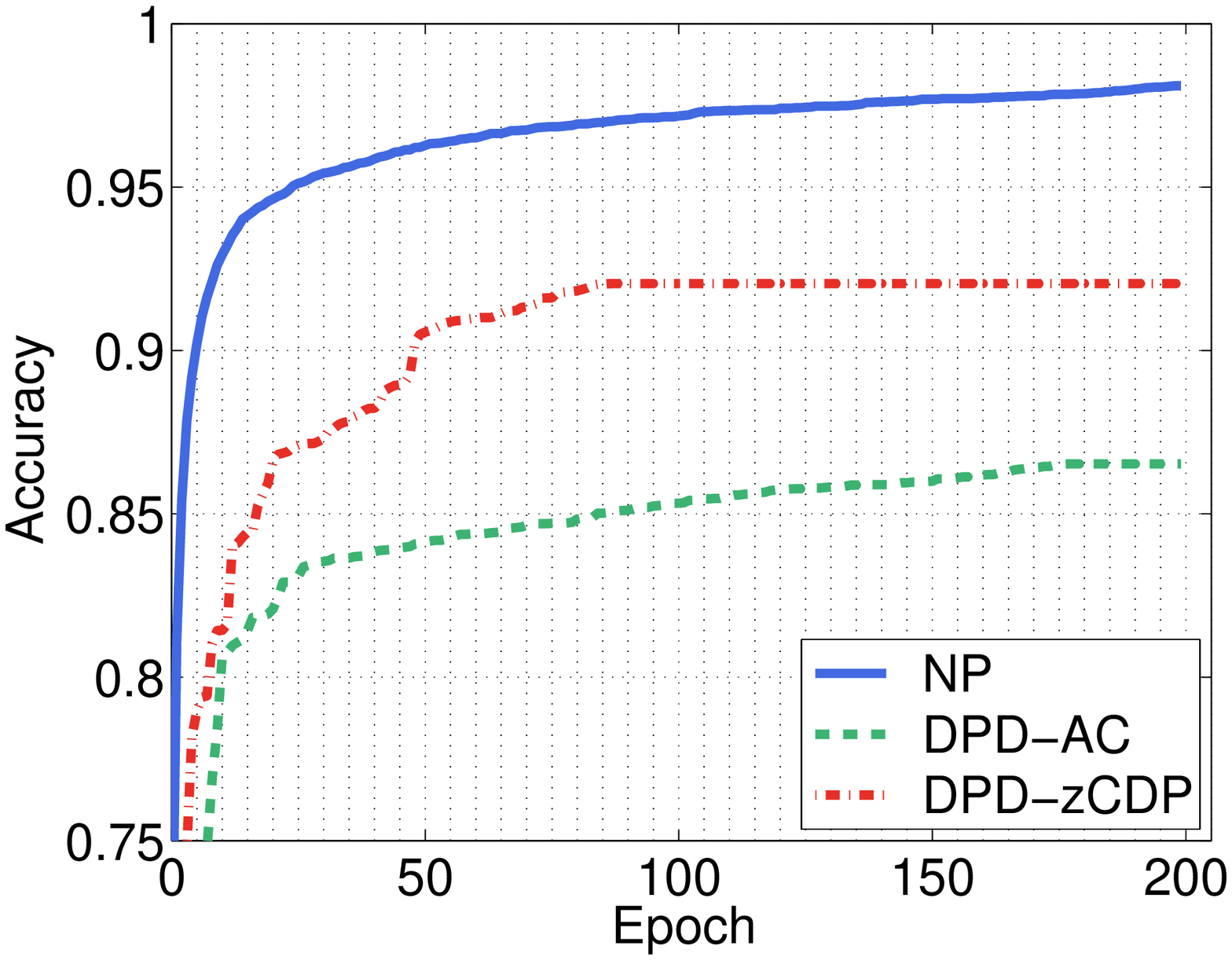} \label{subfig:comp3mnist}}
\end{minipage}
\hspace*{1mm}
\begin{minipage}[b]{0.33\textwidth} 
\centering
\subfigure[DIGITS]{\includegraphics[scale=0.25]{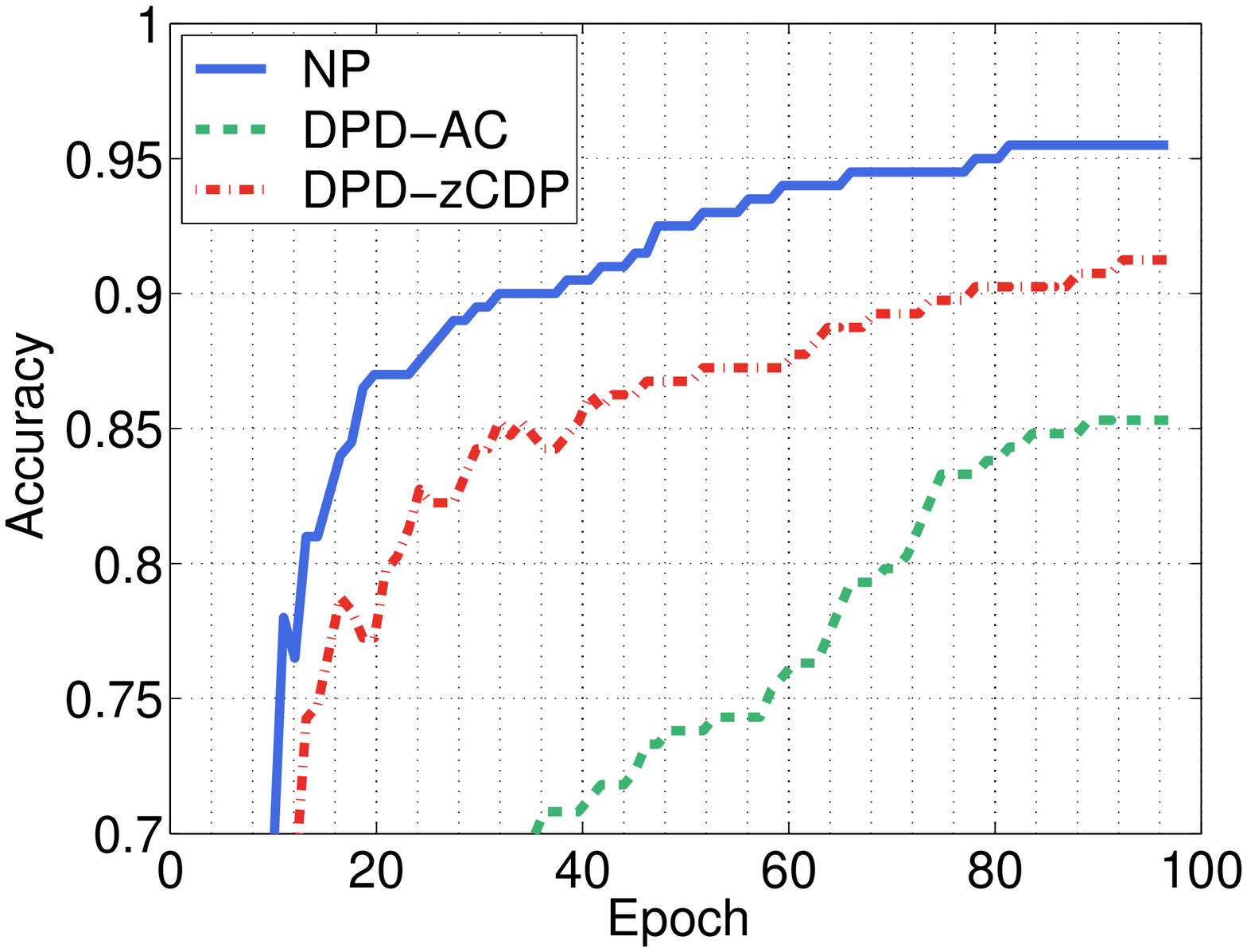} \label{subfig:comp3digits}}
\end{minipage}
\hspace*{1mm}
\begin{minipage}[b]{0.33\textwidth}
\centering
\subfigure[Number of hidden units]{\includegraphics[scale=0.25]{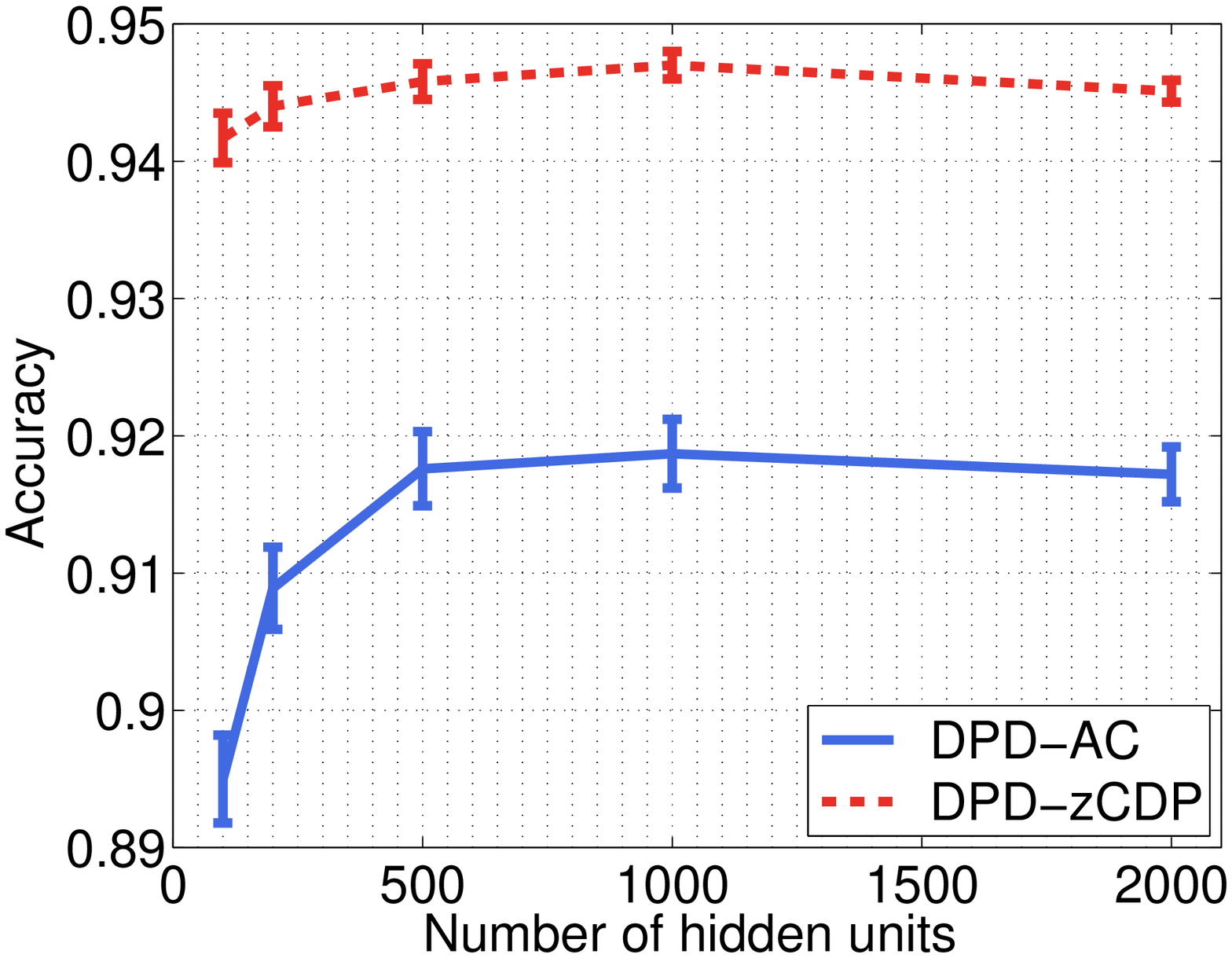} \label{subfig:compHiddenU}}
\end{minipage}   
\\
\hspace*{-5mm}
\begin{minipage}[b]{0.33\textwidth}
\centering
\subfigure[Number of epochs]{\includegraphics[scale=0.25]{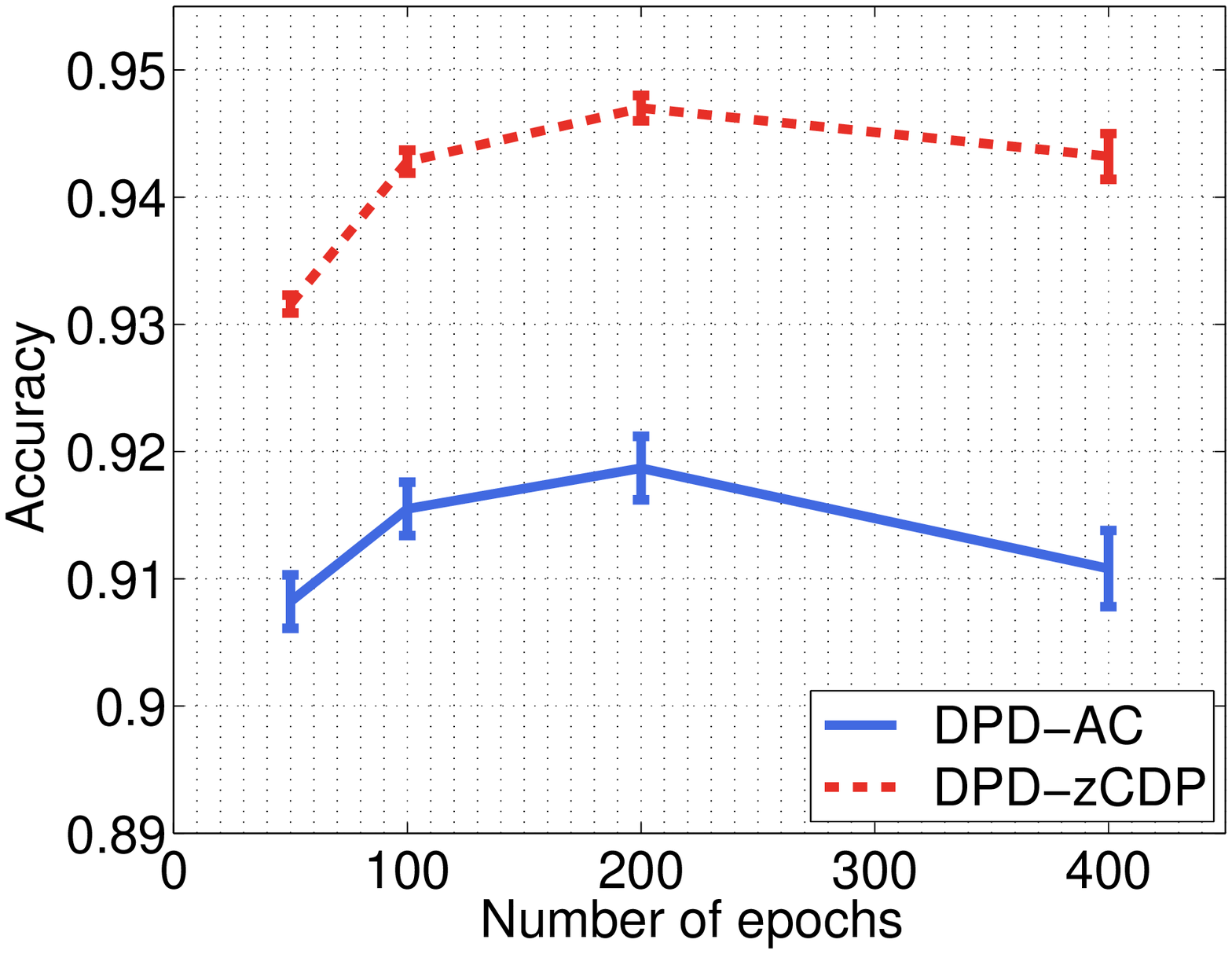} \label{subfig:compEpoch}}
\end{minipage}
\hspace*{1mm}
\begin{minipage}[b]{0.33\textwidth} 
\centering
\subfigure[Gradient clipping norm]{\includegraphics[scale=0.25]{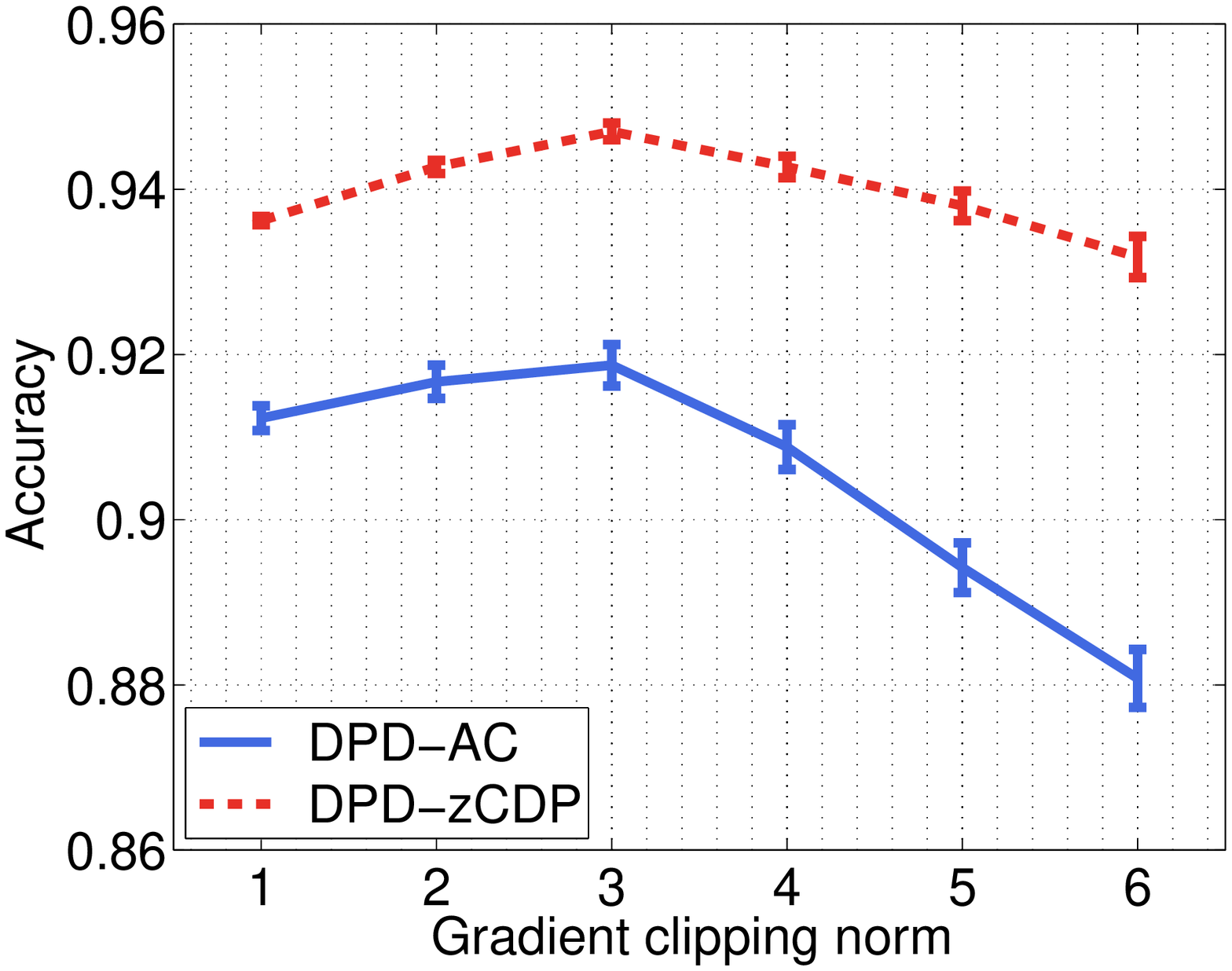}  \label{subfig:compGradClip}}
\end{minipage}
\hspace*{1mm}
\begin{minipage}[b]{0.33\textwidth} 
\centering
\subfigure[Minibatch size]{\includegraphics[scale=0.25]{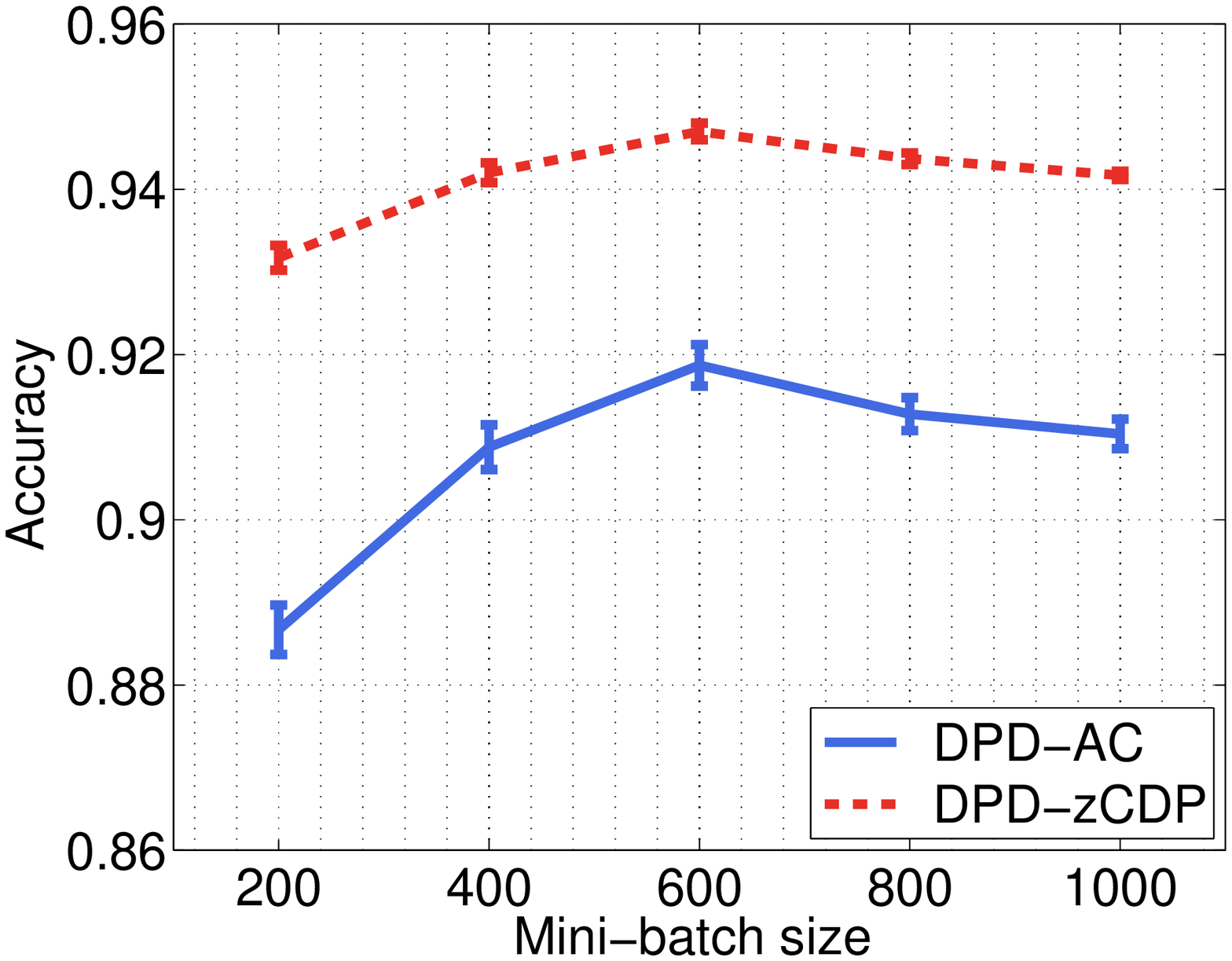} \label{subfig:compMbSize}}
\end{minipage}
\caption{(a) and (b) are the test accuracy results of DPD-AC and DPD-zCDP for $\epsilon$ = 0.5 on MNIST and DIGITS. (c), (d), (e) and (f) show the effect of the model parameters on MNIST dataset.}
\label{fig:MnistPar}
\end{figure*}

Then, we compare the classification accuracy of models learned using two variants of our algorithm: DPD-AC and DPD-zCDP. As mentioned in Section~\ref{sec:dp_budget}, zCDP composition provides a tighter bound on the privacy loss compared to the advanced composition theorem. 
Here we compare them using some concrete values. The noise level can be computed from the overall privacy loss $\epsilon$, the sampling ratio of each minibatch $\nu$ = $S/N$ and the number of epochs $E$ (so the number of iterations is $T$ = $E/\nu$). For our MNIST and DIGITS experiments, we set $\nu$ = 0.01, $E$ = 200 and $\nu$ = 0.05, $E$ = 100, respectively. Then, we compute the value of $\sigma$. 
For example, when $\epsilon$ = 0.5, the $\sigma$ values are 10.88 for DPD-AC and 3.23 for DPD-zCDP on MNIST. 
We can see from Figure~\ref{subfig:compSigmaM} and Figure~\ref{subfig:compSigmaD} that we get a much lower noise by using the zCDP for a fixed the privacy loss $\epsilon$. 
Therefore, for our models with a total privacy budget fixed to $\epsilon$, the amount of noise added is smaller for zCDP, and the test accuracy is higher. Figure~\ref{subfig:comp3mnist} and  Figure~\ref{subfig:comp3digits} show the comparison results of DPD-AC and DPD-zCDP methods when $\epsilon$ = 0.5 with NP model. Both results clearly show that using the zCDP composition further helps in obtaining even more accurate results at a comparable level of privacy.

Lastly, we compare our methods to the most related algorithm proposed by Abadi \emph{et al.}~\citep{Abadi2016} and the case when no dropout is used. For the algorithm with no dropout, we use SGLD to update the weights of the neural network. We ran all the methods on MNIST and DIGITS with varying $\epsilon$. Table~\ref{table:Comparison} reports the test accuracies of all methods. The previous experiments have already demonstrated that DPD-zCDP significantly improves the prediction accuracy. These results also support it and show that \emph{dropout} improves the prediction accuracy 
especially when the privacy budget is low.
\renewcommand{\arraystretch}{1.3}
\begin{table*}[t]
\caption{Comparison of the methods for $\epsilon = \lbrace 10, 1, 0.5 \rbrace$. Bold values indicate the best results.}
\begin{center}
\scalebox{0.90}{
\begin{tabular}{|  L{45mm}  | C{12mm}  | C{12mm}  | C{12mm} ||  C{12mm}  |  C{12mm}  |  C{12mm} |}
\hline     &   \multicolumn{3}{c ||}{MNIST }   &  \multicolumn{3}{c |}{DIGITS }    \\  \cline{2-7}
              &  $\epsilon$ = 10   &   $\epsilon$ = 1     &   $\epsilon$ = 0.5                  &   $\epsilon$ = 10   &   $\epsilon$ = 1     &   $\epsilon$ = 0.5   \\  \hline
DPD-AC         &  0.9526   &    0.9187                &  0.8658                                 &  0.9341                  &   0.9089                &   0.8521                 \\  \cline{1-7}
DPD-zCDP     &  0.9718    &   \textbf{0.9470}  &  \textbf{0.9205}                   &  \textbf{0.9518}   &   \textbf{0.9367}  &   \textbf{0.9125}   \\  \cline{1-7}
SGLD-zCDP (no dropout)  &  0.9581   &    0.9216   &  0.8952                                 &  0.9403                  &   0.9187                &   0.8821                 \\  \cline{1-7}
Abadi \emph{et al.}~\citep{Abadi2016} & \textbf{0.9720}  &  0.9305  &  0.8965   &  0.9480                  &   0.9265               &   0.9003                 \\ \hline
\end{tabular}}
\label{table:Comparison}
\end{center}
\end{table*}

\paragraph{\textbf{Effect of the parameters: }}
The classification accuracy in neural networks depends on a number of factors that must be carefully tuned to optimize the performance. For our DP methods, these factors include the number of hidden units, the number of iterations, the gradient clipping threshold, the minibatch size and the noise level. In the previous section, we compared the effect of different noise levels to the classification accuracy. Here, we demonstrate the effects of the remaining parameters. 
We control a parameter individually by fixing the rest as constant. For MNIST experiments, we set the parameters as follows: 1000 hidden units, minibatch size of 600, gradient norm bound of 3, initial learning rate of 0.1, 200 epochs and the privacy budget $\epsilon$ to 1. The results are presented in Figure~\ref{fig:MnistPar}. 

In non-private neural networks, using more hidden units is often preferable and increases the prediction accuracy of the trained model. For differentially private training, using more hidden units leads more noise added at each update due to the increase in the sensitivity of the gradient. However, increasing the number of hidden units does not always decrease accuracy since larger networks are more tolerant to noise. Figure~\ref{subfig:compHiddenU} shows that accuracy is very close for a hidden unit number in the range of $\left[500, 2000\right]$ and peaks at 1000. 

The number of epochs $E$ (so the number of iterations is $T$ = $E/\nu$) needs to be sufficient but not too large. The privacy cost in zCDP increases in proportion to $\sqrt{T}$ and it is more tolerable than DP. We tried several values in the range of $\left[50, 400\right]$ and observed that we obtained the best results when $E$ is between 100 and 200 for MNIST.

Tuning the gradient clipping threshold depends on the details of the model. If the threshold is too small, the clipped gradient may point in a very different direction from the true gradient. Besides, when we increase the threshold, we add a large amount of noise to the gradients. 
In our experiments, we tried $C$ values in the range of $\left[1, 6\right]$. Figure~\ref{subfig:compGradClip} shows that our model is tolerable to the noise up to $C$ = 4, then the accuracy decreases marginally.

Finally, we monitor the effect of the minibatch size. In DP settings, choosing smaller minibatch size leads running more epochs, however, the added noise has a smaller relative effect for a larger minibatch. Figure~\ref{subfig:compMbSize} shows that relatively larger minibatch sizes give better results. Empirically, we obtain the best accuracy when $S$ is around 600 (so $\nu = S/N = 0.01$). Due to space limitations, we only present the results on MNIST data, the results on DIGITS data have very similar behavior.

\section{Conclusion}
\label{sec:conc}

We introduced differentially private dropout (DPD) that outputs privatized results in deep neural networks with accuracy close to the non-private model learning results, especially under reasonably strong privacy guarantees. To make effective use of the privacy budget over multiple iterations, we proposed to calculate the cumulative privacy cost by using zCDP. Then, we showed how to perform DPD in private neural network training settings and illustrated the effectiveness of our algorithm on several benchmark datasets for DNNs. 

One natural next step is to extend the approach to the distributed training of deep neural networks. The algorithm proposed in the paper are generic and it can be applied to any neural network model. We left its application to other variants of neural networks such as convolutional and recurrent neural networks for future work.




\bibliography{sample_paper}
\bibliographystyle{plainnat}

\renewcommand{\bibname}{}

\end{document}